\PassOptionsToPackage{numbers, compress}{natbib}

\documentclass{article}

\usepackage[main, final]{neurips_2026}
\makeatletter
\renewcommand{\@noticestring}{Accepted as an oral presentation at the AI Agents for Discovery in the Wild (AID-Wild), Workshop at ACM CAIS 2026.}
\makeatother

\usepackage{amsmath,amssymb,amsfonts}
\usepackage{graphicx}
\usepackage{xcolor}

\usepackage{xspace}
\usepackage{booktabs}
\usepackage{caption}
\usepackage{subcaption}
\usepackage{enumitem}
\usepackage{algorithm}
\usepackage{algpseudocode}
\usepackage{listings}
\usepackage{textcomp}
\usepackage[hyphens]{url}
\usepackage{hyperref}
\usepackage{cleveref}
\usepackage[table]{xcolor}
\usepackage{booktabs}
\usepackage{tabularx}
\usepackage{graphicx}
\usepackage{tikz}

\def\BibTeX{{\rm B\kern-.05em{\sc i\kern-.025em b}\kern-.08em
    T\kern-.1667em\lower.7ex\hbox{E}\kern-.125emX}}

\newcommand{\ours}{AI-\textsc{Propeller}\xspace}

\newcommand{\ayazdan}[1]{}
\newcommand{\mamathaananda}[1]{}

\title{\ours: Warehouse-Scale Interprocedural Code Layout Optimization with AlphaEvolve}

\author{%
\textbf{Chaitanya Mamatha Ananda}$^{\ast}$\textsuperscript{\nospacegooglelogo\ucrlogo}\quad
\textbf{Rajiv Gupta}\textsuperscript{\ucrlogo}\quad
\textbf{Mircea Trofin}\textsuperscript{\nospacegooglelogo} \quad
\textbf{Aiden Grossman}\textsuperscript{\nospacegooglelogo} \\
\textbf{Sriraman Tallam}\textsuperscript{\nospacegooglelogo} \quad
\textbf{Xinliang David Li}\textsuperscript{\nospacegooglelogo} \quad
\textbf{Amir Yazdanbakhsh}\textsuperscript{\gdmlogo} \\
University of California, Riverside\textsuperscript{\ucrlogo} \quad
Google\textsuperscript{\nospacegooglelogo} \quad
Google DeepMind\textsuperscript{\gdmlogo}
\\
\footnotesize{\texttt{{\{\href{mailto:cmama002@ucr.edu}{cmama002},\,\href{mailto:rajivg@ucr.edu}{rajivg}\}@ucr.edu}}}\,,\footnotesize{\texttt{\{\href{mailto:mtrofin@google.com}{mtrofin},\,\href{mailto:aidengrossman@google.com}{aidengrossman},\,\href{mailto:tmsriram@google.com}{tmsriram},\,\href{mailto:davidxl@google.com}{davidxl},\,\href{mailto:ayazdan@google.com}{ayazdan}\}@google.com}}
}

\begin{document}

\begin{tikzpicture}[remember picture,overlay]
    \node[anchor=north west, xshift=1.5cm, yshift=-3.3cm] at (current page.north west) {
        \includegraphics[width=2.2cm]{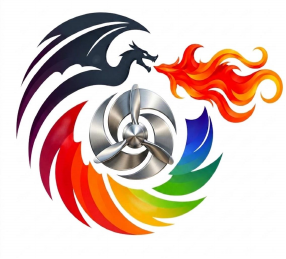}
    };
\end{tikzpicture}

\newcommand{\nospacegooglelogo}{\includegraphics[height=0.8em]{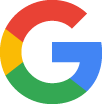}}
\newcommand{\gdmlogo}{\hspace{0.8pt}\includegraphics[height=0.8em]{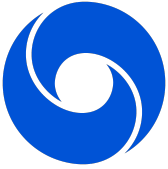}}
\newcommand{\ucrlogo}{\hspace{0.8pt}\includegraphics[height=0.8em]{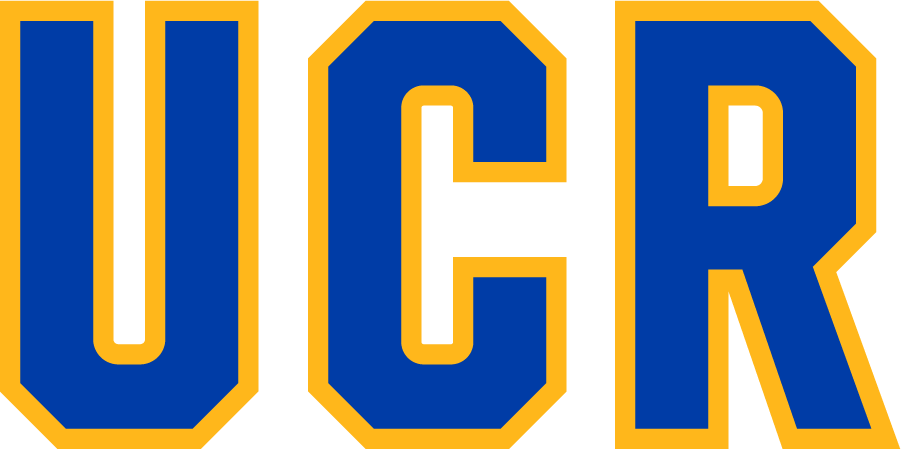}}

\newcommand{\niparagraph}[1]{\smallskip\noindent\textbf{#1.}}
\newcommand{\ninparagraph}[1]{\smallskip\noindent\textbf{#1}}

\definecolor{tableheader}{gray}{0.15} 
\definecolor{tablerow}{gray}{0.95} 

\maketitle
\def\thefootnote{$\ast$}\footnotetext{Work done at Google.}
\setcounter{footnote}{0}
\renewcommand{\thefootnote}{\arabic{footnote}}
\begin{abstract}
Post-link optimizers (PLOs) such as Propeller and BOLT have demonstrated that precise, profile-guided code layout can extract significant performance gains from heavily optimized binaries. 
However, these systems are currently restricted to intraprocedural techniques, leaving the global potential of interprocedural layout largely untapped. 
Interprocedural code layout is historically difficult due to a combinatorially intractable search space and complex call-return semantics that are challenging to model. Consequently, the performance potential of fine-grained interprocedural layout remains unproven in practice.
\ours uses Magellan, an agentic workflow that evolves the compiler heuristic in Propeller into a fine-grained interprocedural optimizer and fine-tunes the resulting policy hyperparameters. 
To ensure high-fidelity, we move away from approximate static cost models and the agentic workflow generates multiple layout variants that are executed on actual hardware to measure real performance counters, providing a precise reward signal for the evolutionary loop. 
\ours has been evaluated on several benchmarks including large warehouse-scale applications and experiments show performance improvements of $0.23$\% to $1.6$\% optimized with state-of-the-art FDO and PLO which is significant for real-world binaries. 
This is the first time ever that large warehouse-scale applications in industrial settings have been optimized with fine-grained interprocedural code layout.
\end{abstract}
\section{Introduction}

\begin{figure*}[!t]
\centering
\includegraphics[width=1.0\linewidth]{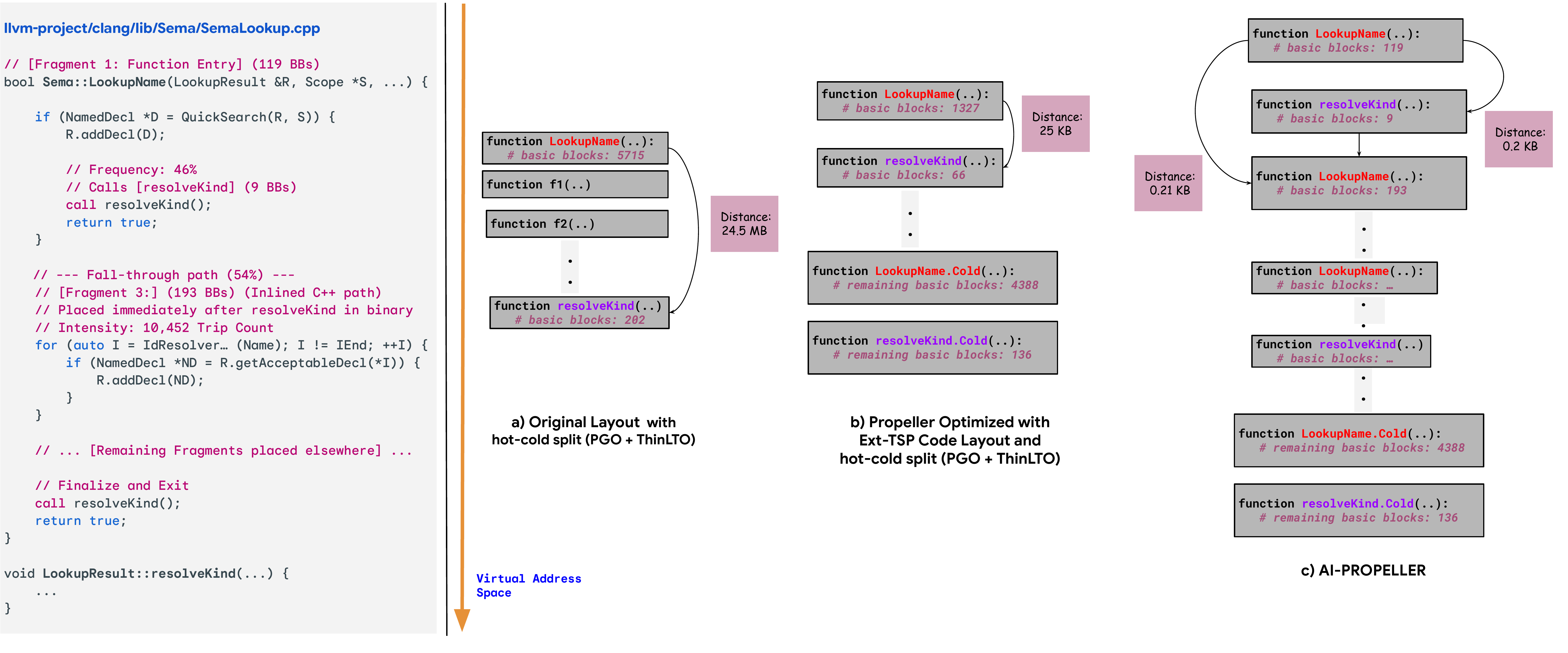}
\caption{Motivating Example from a real-world application, {\em LLVM Clang} binary optimized with \ours}
\label{fig:MotivatingExample}
\end{figure*}

Post-Link Optimizers (PLO) such as Propeller~\citep{propeller} and BOLT~\cite{bolt} have been used to significantly improve the performance of large warehouse-scale applications~\cite{warehouse_scale}. Recently, systems such as ACE~\cite{ace} have demonstrated the potential for AI-driven optimization in warehouse-scale environments, though interprocedural layout optimizations remain an open challenge. At this scale, even marginal performance improvements~\cite{asmdb,temeriare} have significant business value. Code Layout~\cite{pettis,exttsp} is an important optimization that has been particularly effective for such applications. These PLO optimizers use accurate profiles to reorder the basic blocks of hot functions, optimizing instruction cache and TLB utilization. Further, function splitting~\cite{funcsplitting1990, mfs} cleaves the frequently executed (hot) and rarely executed (cold) subsets of a function and places them in separate regions in memory to further improve TLB behavior. Collectively, these optimizations squeeze maximal performance by placing the hot working set of an application to reside in close proximity.

Code layout has been well studied and a number of prior works exist that reorder both basic blocks~\cite{ICBE,program_opt_for_icache,conditionalbr_code_replication,Ispike,codestitcher} and functions~\cite{pettis,mcfarlingProcMerge,GloyProcPlacement} for optimal utilization of instruction caches and TLBs. Early research~\cite{Chang1991} demonstrated the value of profile information in guiding code optimizations, while techniques such as trace scheduling~\cite{tracescheduling} pioneered global code optimization to maximize instruction-level parallelism. Particularly, the Ext-TSP heuristic~\cite{exttsp} has proven to be very effective and is used by both Propeller and BOLT.  However, these heuristics are applied intraprocedurally, one function at a time, and do not cross function boundaries with the exception of function splitting. While PLO frameworks such as Propeller possess the capabilities to reorder basic blocks across procedures arbitrarily (e.g. via basic block sections~\cite{propeller}), there are no existing code layout heuristics that exploit this opportunity to reorder blocks across functions for increased performance.  

Interprocedural code layout, that is the ability to partition functions and reorder the partitions, is a significantly more challenging problem in comparison to intraprocedural layout, for two main reasons.  First, the combinatorial complexity grows exponentially as the interprocedural search space is several orders of magnitude larger. For example, a warehouse-scale application we optimized has $\sim1.4$ million functions and the largest function has $\sim20,000$ basic blocks.  Whereas, the entire application has $\sim40$ million basic blocks which adds several million orders of complexity over intraprocedural exploration. Scaling to an interprocedural algorithm is a significant engineering effort, which is hard to justify without demonstrable performance improvements. Second, the cost function to layout basic blocks across procedures is very different as call-return semantics cannot be violated even if two blocks from different procedures are placed adjacently, whereas, with intraprocedural layout the fall-through jumps between adjacent blocks are eliminated reducing the dynamic instruction count and also increasing cache utilization. To our knowledge, \ours is the first work to demonstrate clear performance potential on large warehouse-scale applications via interprocedural code layout.

\ours adopts the Magellan framework \cite{magellan} to evolve code heuristics. Specifically, it leverages {\em AlphaEvolve}~\cite{alphaevolve} to evolve existing heuristics, and a black-box optimizer, such as Vizier \cite{vizier} to tune its numerical parameters. Both AlphaEvolve and Vizier rely on a \textit{reward signal} to propose code edits and parameter updates, respectively. The reward is expected to be empirically obtained by applying the new heuristics to binaries of interest (or - for completeness - closely correlate with an empirical signal of this nature). More details in section \ref{sec:magellan}.

In this context, we apply the framework to Propeller's \cite{propeller} code layout optimization, which is currently formulated as an {\em Extended Traveling Salesman Problem}~\cite{exttsp}.
The key challenge that had to be overcome when applying Magellan was ensuring the reward signal is both \textit{time -- efficient} (to maintain practical training times) and \textit{noise-free} (to prevent misleading AlphaEvolve or Vizier). To that end, we identified a small set of representative benchmarks (binaries comparable in working set size to the binaries of interest) that had a very low run-to-run variation in performance (lower than $0.05\%$). 
This paper makes the following contributions:

\begin{enumerate}[leftmargin=*]
\item
This is the first study to show performance improvements with interprocedural code layout optimizations on real-world applications, including large warehouse-scale binaries, resulting in performance improvements ranging from $0.23\%$ to $1.6\%$. We want to stress that, while naively these improvements may seem small, they are both significant from a operational cost / business perspective, and very hard to achieve in industrial settings already employing the state of the art optimization techniques (profile-guided optimizations, link-time optimizations, and the current block-layout techniques)
\item
We show that interprocedural block placement heuristics are achievable for large binaries in industrial settings without noticeably changing build characteristics (time and memory usage)

\end{enumerate}

\section{Interprocedural Code Layout in Propeller}
\label{sec:propeller_writeup}

Propeller~\cite{propeller} is an industry-strength Post-Link Optimizer that has been used to optimize large data-center applications, particularly in environments with distributed build systems where build scalability is critical. As noted in the paper~\cite{propeller}, the improved performance is primarily from intraprocedural code layout and function splitting~\cite{mfs}. Intraprocedural code layout optimizes each function's basic block layout for locality and reduces the number of dynamic taken-branches by placing two hot basic blocks that are successively executed adjacent to each other.  Function splitting separates the hot and cold subsets of each function and clusters the hot and cold parts of all functions into separate regions.  This reduces the working set size and improves locality and performance.

Propeller introduced {\em basic block sections} which allows arbitrary code layout of basic blocks in the virtual address space, even across functions.  While Propeller has the capabilities for interprocedural code layout, it currently does not optimize binaries with interprocedural heuristics.   As noted in the Propeller paper~\cite{propeller} in {\em Section 4.7}, while one isolated experiment with interprocedural layout did show a small performance improvement on one benchmark, generating such layouts took a very long time ($3X$ to $10X$) primarily due to the increased search space, and interprocedural layout has been labeled as future work.

\ours has been implemented in the Propeller framework. As it is already able to lay out code interprocedurally, the challenge was to find a heuristic that yields performance improvements above and beyond the state-of-the-art.  We used {\em Magellan} to evolve the existing heuristic resulting in an improved interprocedural capable heuristic without affecting the build scalability.  We have used \ours to optimize large and warehouse-scale applications without any perceivable degradation in build characteristics such as time and memory usage.

\section{Motivating Example}
\label{sec:motivating_example}
\begin{figure*}[t]
\centering
\includegraphics[width=0.99\linewidth]{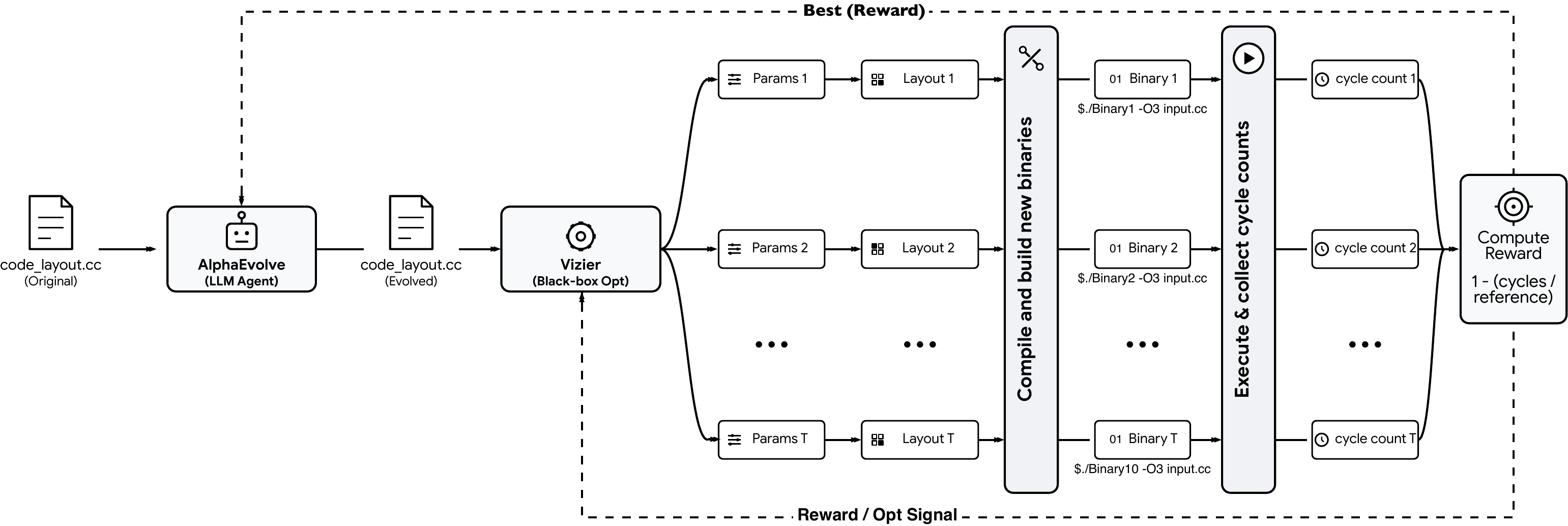}
\caption{\ours Overview: Application of Magellan to Propeller}
\label{fig:overview}
\end{figure*}
To illustrate how interprocedural layout is able to further improve performance, we show a code snippet and the optimized code layout generated by \ours on a real-world application, {\em LLVM Clang}~\cite{ClangCompiler} compiler.  In Figure~\ref{fig:MotivatingExample}, function \texttt{Sema::LookupName} calls \texttt{LookupResult::resolveKind} twice.  In the original layout shown in Figure~\ref{fig:MotivatingExample}(a) for a PGO+ThinLTO~\cite{thinLTO} binary, these functions are placed arbitrarily and far apart, $\sim 24$ MB, in the address space even though the calls are hot.  With Propeller's intraprocedural layout and function reordering, these functions are split into their hot and cold subsets first and their hot subsets are placed close together, $25$ KB apart, and this significantly improves locality resulting in better performance.  The hot function size of \texttt{LookupName} is still $25$ KB even after splitting, due to aggressive inlining. \ours takes this even further.  It identifies the basic block subsets of the hot function partitions that are executed during calls to function \texttt{LookupResult::resolveKind} and places them together, $0.2$ KB apart, significantly improving locality. Traditional hot-cold function splitting~\cite{mfs} only splits a function in two subsets, \ours can split a function into several subsets.

\begin{figure}[t]
    \centering
    \begin{minipage}{0.48\textwidth}
        \centering
        \includegraphics[width=\linewidth]{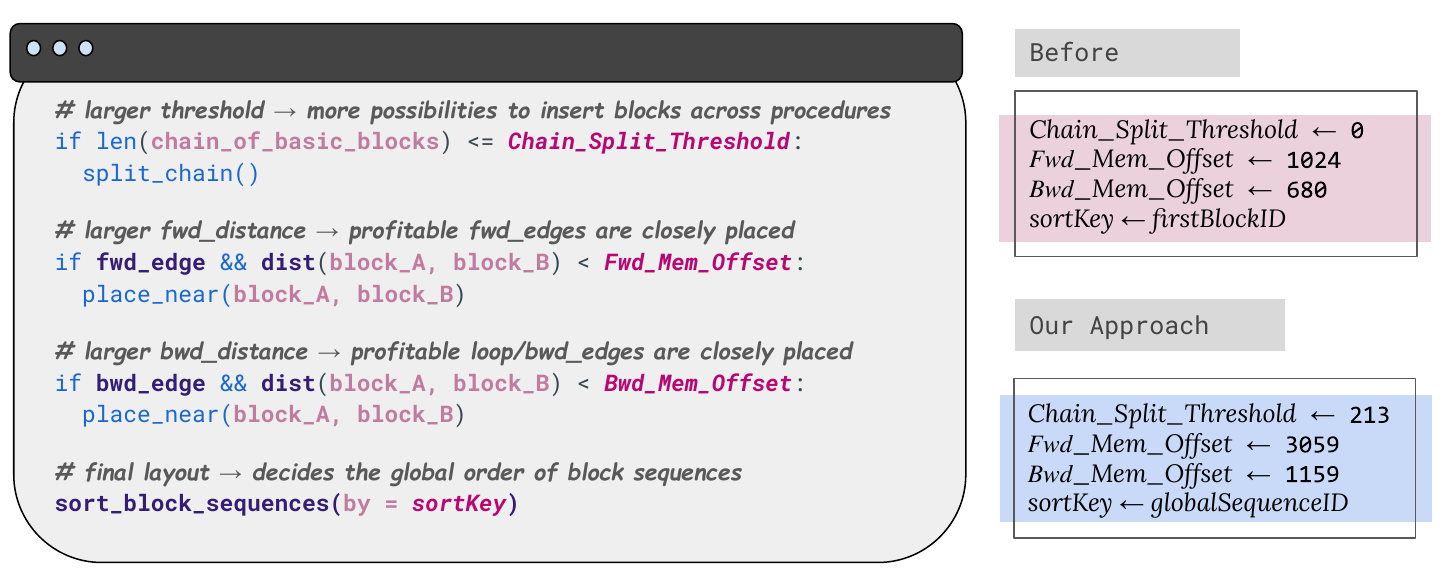}
        \caption{Discovered Policy from \ours for {\em LLVM Clang}~\cite{ClangCompiler}}
        \label{fig:DiscoveredPolicy}
    \end{minipage}\hfill
    \begin{minipage}{0.48\textwidth}
        \centering
        \includegraphics[width=\linewidth]{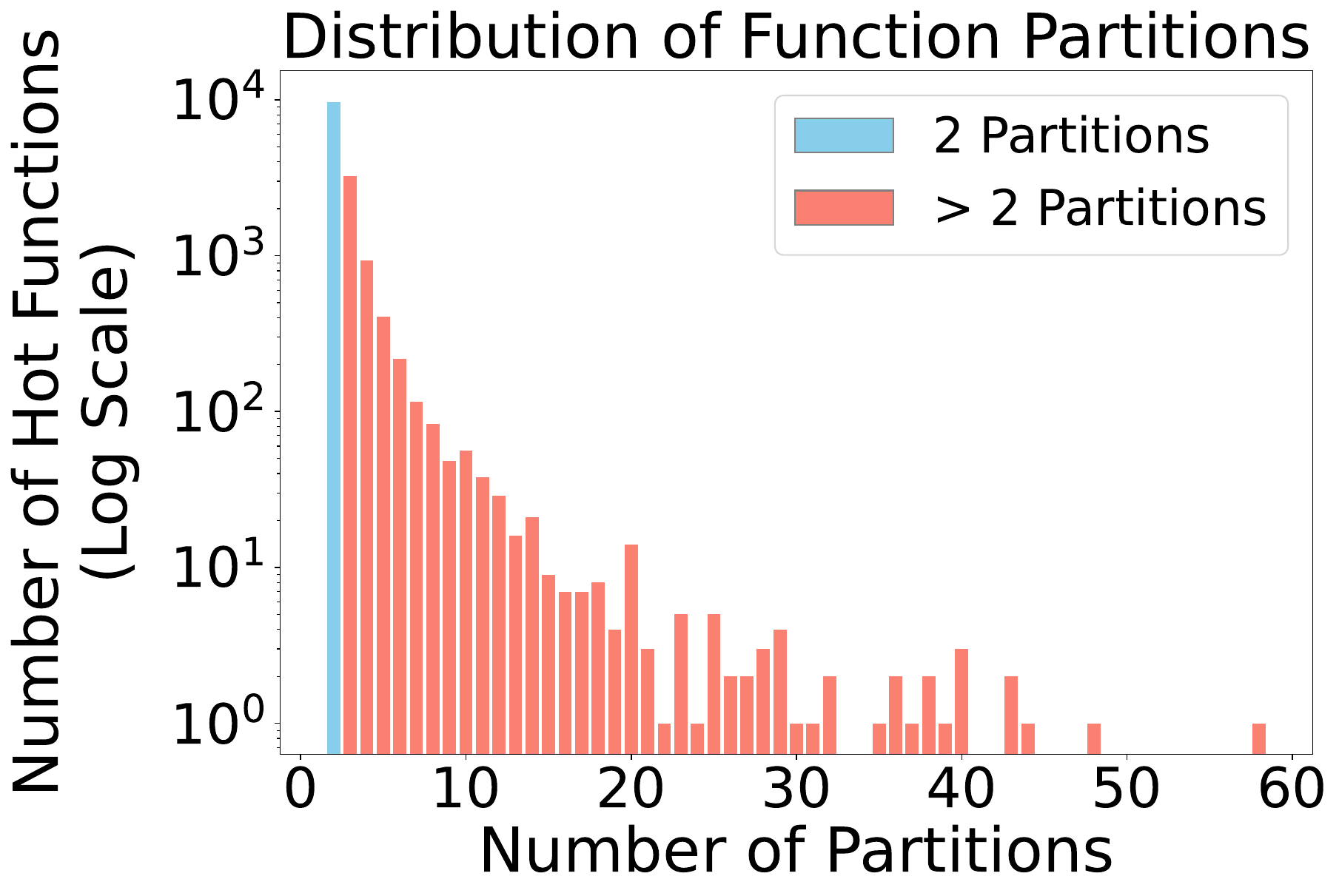}
        \caption{Distribution of Partitions in {\em LLVM Clang}~\cite{ClangCompiler} Binary Optimized with \ours}
        \label{fig:ClangPartition}
    \end{minipage}
    \vspace{-0.2in}
\end{figure}

Looking at the final optimized binary of {\em LLVM Clang}~\cite{ClangCompiler} with \ours, as shown in Figure~\ref{fig:ClangPartition} we found that out of the $\sim15000$ hot functions that  were considered as candidates for interprocedural layout, $\sim5000$ functions were partitioned in more than $2$ ways with the maximum number of partitions being $57$ and the mean number of partitions slightly larger than $3$.  This illustrates both the vastness of the search space with interprocedural layout and the effectiveness of \ours in finding layouts that improve performance.
\section{Overview of \ours using Magellan}
\label{sec:overview}
We will first describe Magellan~\cite{magellan}. Its use in our specific context is depicted in ~\Cref{fig:overview}. Next we discuss the improvements it made to an initially naive inter-procedural block placement heuristic.

\niparagraph{Magellan}\label{sec:magellan}
We use the Magellan approach~\cite{magellan} combining AI-guided heuristic discovery and black-box numerical optimization. This is a reward guided evolution mechanism. The process begins with an initial policy - a function - being presented to AlphaEvolve (AE), together with a LLM prompt. AE prompts the LLM with that prompt and the referenced function and obtains one or more edits on that initial function. Each edit consists of novel code additions / modifications / removals. Since the original function operated on internal Propeller data structures, AE may extract new features of code IR, for example, to consider in making decisions. As part of the Magellan methodology, the prompt includes a directive that any numerical parameters be exposed as compiler flags. 
For a policy thus proposed by AE, Magellan attempts the following:
\begin{enumerate}[leftmargin=*]
    \item Recompile Propeller by replacing the policy being evolved with the proposed policy, in the Propeller codebase.

    If this step fails (because the proposed edits cause compilation failure), Magellan feeds the build error messages back to AlphaEvolve (which includes them in the next prompt) and repeats the process;
    \item For policies that can successfully be included in the Propeller codebase, Magellan identifies their parameters (the generated compiler flags) and starts a blackbox tuning session with e.g. Vizier \cite{vizier}. 
\end{enumerate}
 Vizier attempts to tune those parameters based on the reward signal, which, as discussed, it obtains by running representative binaries in a noise-free environment. The best performing flag combination is presented back to Magellan, which records it and presents it back to AlphaEvolve together with the observed reward. The process then continues - AlphaEvolve samples out of its previously produced policies, mutates the samples, and generates new policies. The overall process has no intrinsic termination point.
\subsection{Evolving the Code Layout Policy in Propeller}
Interprocedural basic block reordering for a binary is governed by three factors. How sequences of basic blocks are split, where new blocks are inserted, and which edges are prioritized for profitability: \textit{Chain Split Threshold}, \textit{Forward Memory Offset}, and \textit{Backward Memory Offset}. The search space has remained unexplored because manual tuning of these interdependent parameters is computationally intractable is of the order of \textit{Billions} for large binaries. By leveraging {\em Magellan}, \ours identified a novel configuration that produces performance improving interprocedural layouts as shown in Figure~\ref{fig:DiscoveredPolicy}.  Specifically, \ours evolved the values of the three constants discussed above that determine the order of the layout. The specific improvements made to the policy are categorized below:

\begin{itemize}[leftmargin=*]
\item \textbf{Optimal Split Points:} The current heuristic forms sequences of basic blocks while optimizing and determines split points in existing sequences to insert new blocks. The \texttt{chain\_split\_threshold} parameter determines the maximum sequence size above which sequences will not be considered for splitting.  The existing Ext-TSP~\cite{exttsp} had this at $0$ avoiding any splitting possibility whereas \ours increased this to $213$, splitting a lot more sequences.

\item \textbf{Considering Edges Over Larger Distances:} The memory offset parameters, \texttt{Fwd\_Mem\_Offset} and \texttt{Bwd\_Mem\_Offset}, determine the maximum distance between two basic blocks for their connecting edge to be considered by the optimizer. The current heuristic~\cite{exttsp} restricts this to 1024 bytes (forward) and 680 bytes (backward). This narrow window is sufficient for \textit{intraprocedural} layout where hot paths are local (confined inside a single function). However, \textit{interprocedural} layouts involve caller-callee and return relationships that frequently span over much larger memory distances. \ours expanded these limits to 3059 and 1159 bytes, respectively, allowing the optimizer to capture and minimize the distance of these previously ignored \textit{long-range} edges.

\item \textbf{Preserving the Interprocedural Layout:} While the numerical parameters determine how the various basic blocks are grouped into sequences, \ours also evolved the logic for the order in which these sequences are organized in the binary. Ext-TSP sorts the various resulting block sequences by their original block IDs. This is effective for \textit{intraprocedural} layout, where the goal is simply to order blocks within a function. As the scope expands to \textit{interprocedural} layout, \ours evolves the logic to support interprocedural optimization goals. To account for the local scope of block IDs, \ours generates a global layout index. By sorting according to this new index rather than function-specific block IDs, the system ensures that the final binary strictly preserves the intended order for basic blocks.
\end{itemize}

\section{Experiments and Results}
\label{sec:exp}

\niparagraph{Training setup}
\ours framework was trained on a representative subset of 100 modules derived from the {\em Ninja}~\cite{cmake, ninja} build of {\em Clang} compiler. The training phase spanned 2.7 days, comprising 12 AlphaEvolve~\cite{alphaevolve} iterations and a total of 1,200 Vizier~\cite{vizier} trials (distributed as 100 trials per AlphaEvolve iteration).

To maintain a high quality (noise-free) reward signal, all performance evaluations were conducted on Intel Skylake systems within a strictly controlled environment. Specifically, Turbo Boost, SMT and ASLR were disabled to minimize non-deterministic noise. Furthermore, the CPU frequency was locked at 75\% of the peak frequency to eliminate thermal throttling and ensure consistent execution across trials.  Each candidate binary was executed 10 times, yielding a run-to-run variance of 0.02\% to 0.04\%. This high stability enabled \ours to accurately attribute fine-grained performance gains to the proposed layout transformations.

\begin{table}[htbp]
    \centering

    \renewcommand{\arraystretch}{1.6} 
    
    \rowcolors{2}{tablerow}{white} 
    
    \begin{tabularx}{0.95\textwidth}{
        >{\bfseries\color{tableheader}}l
        >{\raggedright\arraybackslash}X 
    }
 
        \rowcolor{tableheader}
        \textcolor{white}{\textbf{Configuration}} & \textcolor{white}{\textbf{Description}} \\
        \midrule
        
        Baseline & The FDO~\cite{FDO} and ThinLTO~\cite{thinLTO} optimized binary with function layout disabled. \\
        CDSort~\cite{llvm_cdsortconfig} & The state of the art function layout (reordering) algorithm applied to Baseline. \\
        Ext-TSP~\cite{exttsp} & The state-of-the-art algorithm for \textit{Intraprocedural} basic block reordering applied to Baseline. \\
        \textit{\ours} & Our Approach applied to Baseline. \\
        
        \bottomrule
    \end{tabularx}
    \caption{Summary of Evaluated Algorithms and Configurations}
    \label{tab:algorithms}
\end{table}

\subsection{Experimental Methodology}
\niparagraph{Benchmarks and baseline algorithms}
Our evaluation covers a diverse set of applications, including {\em LLVM Clang}~\cite{ClangCompiler}, {\em LevelDB}~\cite{GhemawatLevelDB}, {\em Redis}~\cite{SanfilippoRedis}, and Search (a warehouse-scale application).
\ours was evaluated and compared against the algorithms described in~\Cref{tab:algorithms}.

\subsubsection{Policy Generalizability}
\label{sec:policy_generation}
We evaluated the generalizability of the policy discovered by \ours in Figure~\ref{fig:DiscoveredPolicy}. While the policy was trained on a small test workload of only 100 modules, it generalized effectively to the full {\em LLVM Clang} build comprising $\sim3,000$ modules. The same policy also performed well on {\em LevelDB} and {\em Redis} benchmarks. However, it did not generalize well to the warehouse-scale {\em Search} workload, yielding neutral performance. Consequently, we trained a separate policy for Search using its specific test workload and evaluated it accordingly.

\subsection{Experimental Evaluation}
To evaluate the effectiveness of \ours, we conducted experiments focusing on front-end bound~\cite{topdown} workloads, where instruction cache efficiency and branch prediction are critical. 

\niparagraph{Evaluation Environment}
All performance evaluations were executed on dedicated, bare-metal infrastructure (OVH Dedicated Server) using {\em LLVM}\footnote{Commit Hash: bf91a62269964398836544020def699e3f019b9b} with \textit{ThinLTO}~\cite{thinLTO} enabled globally. Benchmarks were executed on an Intel Xeon Silver 4214R system running Arch Linux (Kernel 6.17.5). The hardware of the system consists of 48 cores and 96 GB of RAM. The cache hierarchy includes a 32 KB L1 instruction/data cache, a 1 MB private L2 cache per core, and a 16.5 MB shared L3 cache. 

\subsection{Analysis of Results}

Our results demonstrate that \ours consistently outperforms traditional heuristics by discovering non-intuitive, interprocedural layouts. Across all the benchmarks, the performance gains are statistically significant with a $p$-value close to zero. 

\begin{figure}[htbp]
    \centering

    \begin{minipage}{0.48\textwidth}
        \centering
        
        \includegraphics[width=\linewidth]{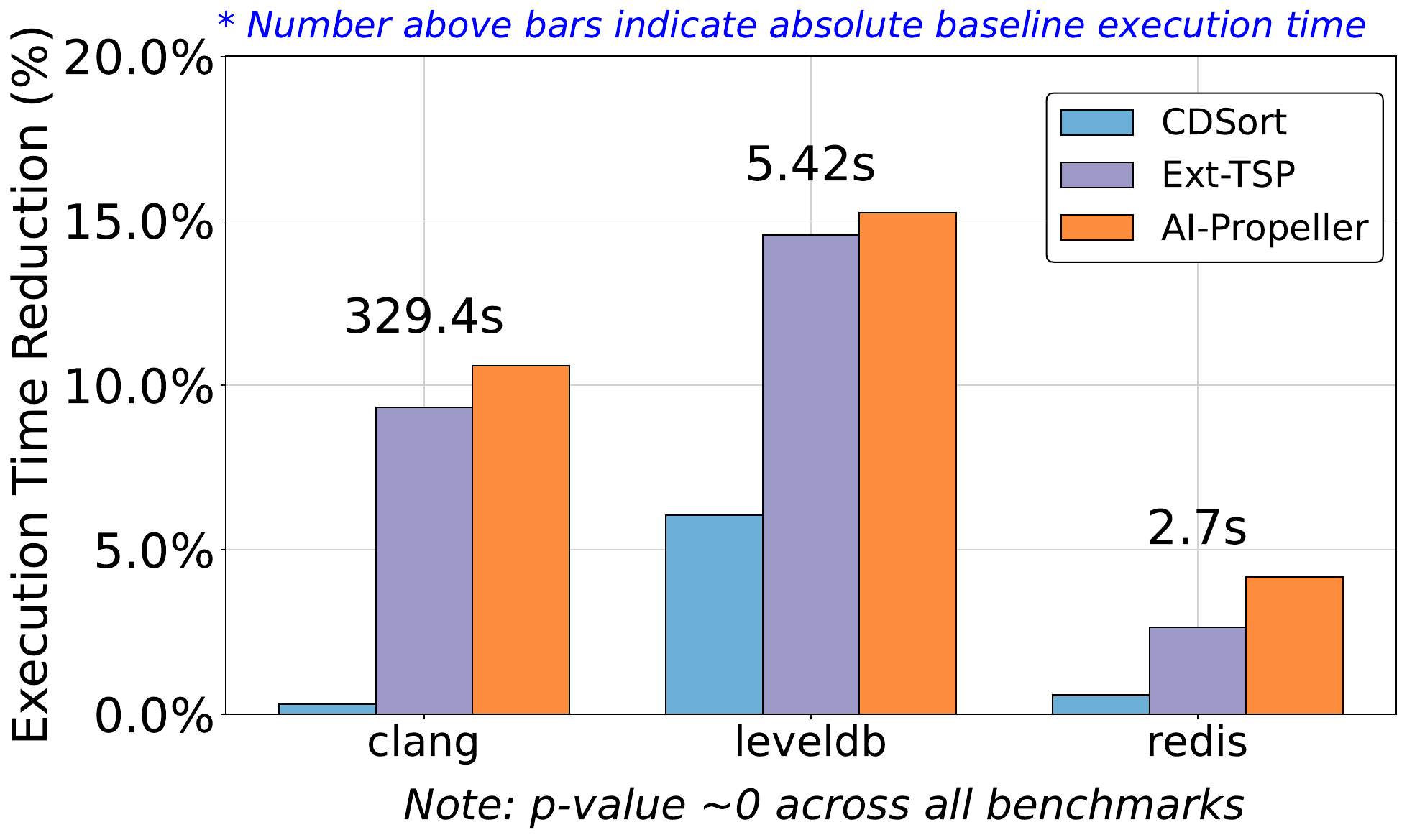}
        \caption{Execution Times of baseline and relative changes in execution times achieved by CDSort, Ext-TSP and \ours.}
        
        \label{fig:performanceImprovement_exectime}
    \end{minipage}\hfill
    \begin{minipage}{0.48\textwidth}
        \centering
        
        \includegraphics[width=\linewidth]{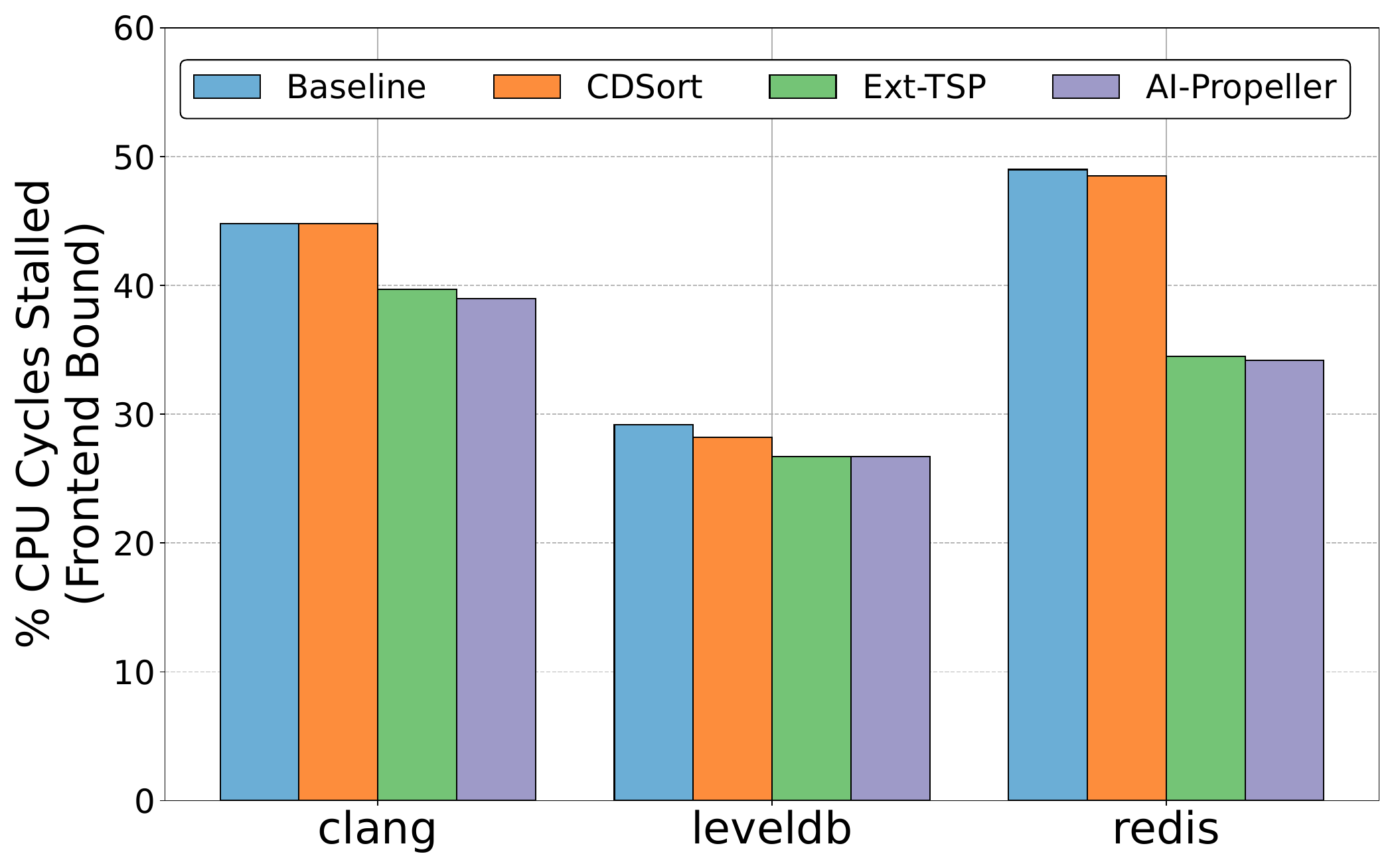}
        \caption{\% of CPU cycles stalled on front-end events~\cite{topdown} across different algorithms, lower is better.}
        \label{fig:performanceImprovement_stalls}
    \end{minipage}
\end{figure}

\niparagraph{Reduction in execution times}
The evaluation, Figure~\ref{fig:performanceImprovement_exectime}, of \ours demonstrates performance gains over the \textit{Baseline} and prior state of the art across three distinct applications.  Particularly, on a large real-world application like {\em clang}, the improvement is $1.6$\%.  To put this in perspective, Propeller~\cite{propeller} alone improved this benchmark by $7.3$\% as noted in the paper. \ours further squeezes more than  20\% of that improvement. 

\niparagraph{Frontend-Bound Stalls~\cite{topdown}}
A common bottleneck across these workloads is high instruction fetch and decode stalls. \ours significantly mitigates these stalls, Figure~\ref{fig:performanceImprovement_stalls} by improving code locality. In {\em Clang}, we observed a reduction in stalls compared to baseline from 45\% to 39\%. In {\em LevelDB} stalls decreased from 29\% to 27\%, while {\em Redis}, frontend-bound stalls dropped from 49\% in the baseline to 34.2\%, showing significant reduction in frontend stalls.

\begin{figure}[t]
    \centering
    \includegraphics[width=0.45\linewidth]{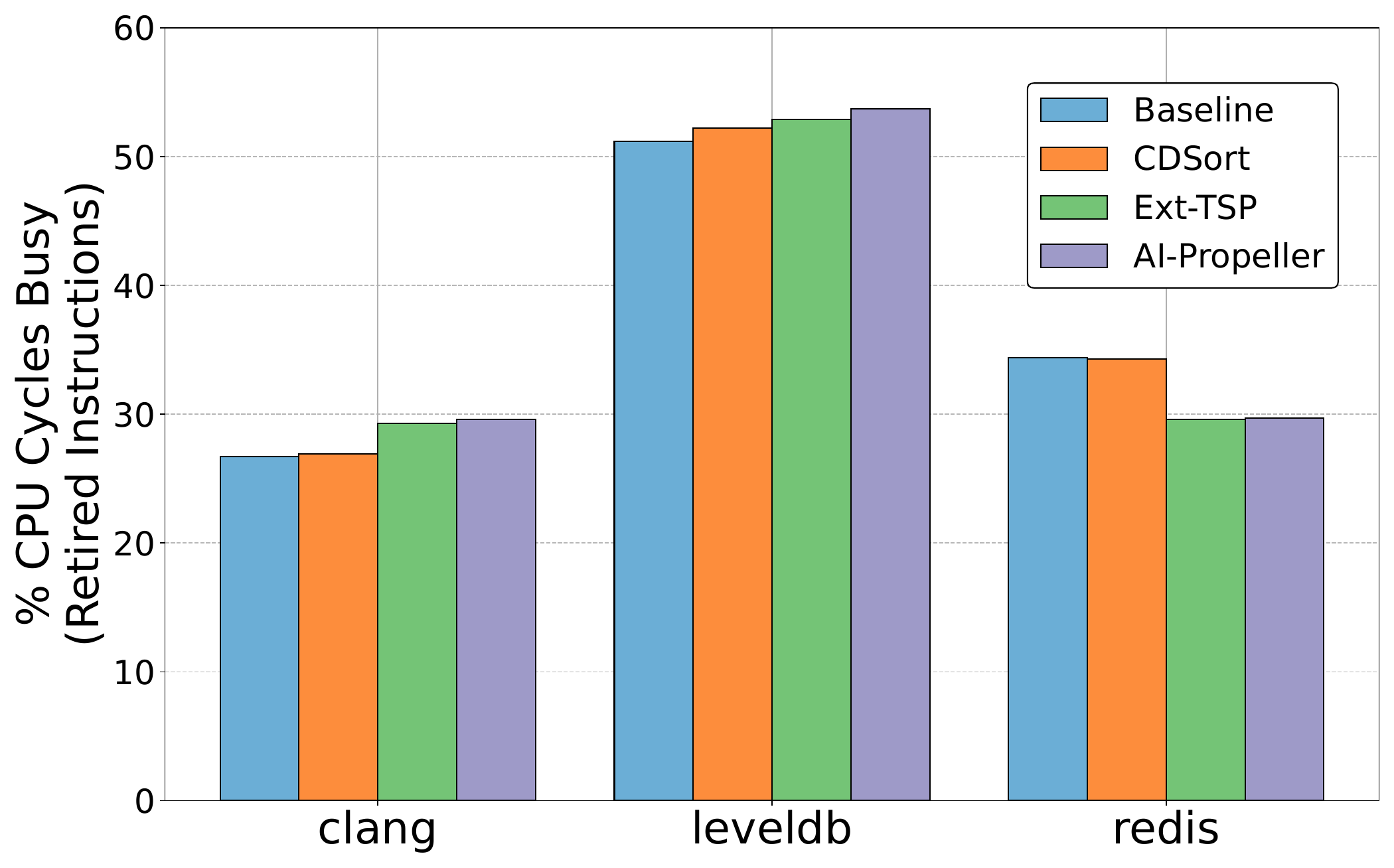}

    \caption{\% of CPU pipeline slots spent in retiring instructions. Higher is better as it indicates more useful work done.}
    \label{fig:performanceImprovement_insns}
\end{figure}

\niparagraph{CPU pipeline slots spent retiring instructions}
With reducing frontend stalls, top-down analysis~\cite{topdown} shows that \ours enables the processor to sustain a higher rate of retired instructions, Figure~\ref{fig:performanceImprovement_insns}, which is a key indicator of useful work performed per cycle. \ours improved instruction retirement throughput in {\em LLVM Clang}~\cite{ClangCompiler} from 27\% to 30\%, in {\em LevelDB}~\cite{GhemawatLevelDB} it improved from 51\% to 54\% and in {\em Redis}~\cite{SanfilippoRedis}, we did not see an increase over the baseline, it remained comparable to Ext-TSP.

\subsection{Warehouse Scale}
\label{sec:warehouse}
To evaluate \textit{\ours}, we applied it to a proprietary warehouse workload, \textit{Search}. For the AlphaEvolve~\cite{alphaevolve} generated policy from clang, Vizier~\cite{vizier} suggested a distinct set of hyper-parameters for \textit{Search}. Unlike clang benchmark, for \textit{Search}, \textit{\ours} converged on to a more conservative chain split threshold of 53, while expanding memory offsets to 2074 (forward) and 1423 (backward) to account for the massive code footprint.

This discovered policy yielded a 0.23\% improvement over the deployed Ext-TSP at production. While proprietary constraints limited the collection of frontend stall and instruction retirement metrics, this result demonstrates that \textit{\ours} can identify non-obvious optimization opportunities even in hyper-optimized, warehouse-scale workloads.

\subsection{Inter vs. Intra}
We conducted an ablation study to isolate the performance gains we obtained on {\em clang} with \ours and identify the portion of the wins that were primarily due to interprocedural layout.

We modified the optimal interprocedural layout generated by \ours and forced all basic blocks belonging to the same function to remain contiguous, effectively disabling interprocedural layout and preserving the intraprocedural orderings.

When these interprocedural transformations were reversed, the performance improvement in cycle counts dropped from 1.6\% to just $\sim {0.3}$\% over Ext-TSP. This significant delta indicates that while \textit{\ours} does identify marginal intraprocedural improvements, the vast majority of the total speedup, approximately 80\% is derived from its ability to interleave blocks across function boundaries. This confirms that global interprocedural reordering is the primary driver of the \ours framework's effectiveness.

Restricting \ours to train on intraprocedural layout of basic blocks for {\em LLVM Clang}, we observed only about $\sim{0.6}$\% improvement over Ext-TSP on {\em clang} reinforcing the need for interprocedural basic block reordering.
\section{Related Work}
\label{sec:related_work}

Prior code layout optimization techniques fall into three broad categories: function-level reordering, basic-block reordering, and learning-based approaches. These techniques rely on static cost models that sometimes do not capture the nuances of the hardware leading to sub-optimal layout decisions.

\niparagraph{Function-Level reordering}
Procedure-level placements improved instruction cache and TLB locality via profile-guided positioning~\cite{pettis}, temporal execution data, and procedure merging~\cite{mcfarlingProcMerge, GloyProcPlacement}. Modern large-scale systems refine this using dynamic clustering heuristics, such as \textit{HFSort} and \textit{C3}~\cite{hfsort_c3}, or cache-aware placement like \textit{CDSort}~\cite{llvm_cdsortconfig}. However, these coarse-grained methods treat functions as atomic units. They cannot separate hot and cold regions or co-locate hot blocks across function boundaries, which are the limitations our fine-grained approach directly resolves.

\niparagraph{Basic block reordering}
Basic-block reordering addresses these atomic limitations~\cite{program_opt_for_icache}, with early work exploring code replication for branch prediction~\cite{conditionalbr_code_replication}. The \emph{Extended Traveling Salesperson based heuristic (Ext-TSP)}~\cite{exttsp} is the state-of-the-art in intraprocedural code layout and utilized by modern post-link optimizers like \textit{BOLT}~\cite{bolt} and \textit{Propeller}~\cite{propeller}. While Propeller's basic block sections and Codestitcher~\cite{codestitcher} enable cross-function block stitching, these frameworks still rely on static objective functions. Further, CodeStitcher~\cite{codestitcher} relies on whole program link time optimizations that do not scale with large distributed build systems. Consequently, they struggle to effectively navigate the interprocedural search space.

\niparagraph{Learning-Based optimization approaches}
Machine learning is used to augment or replace static compiler heuristics. MLGO~\cite{mlgo} applied neural network models to replace heuristics in an industrial setting. PIE~\cite{pie1, pie2} utilizes LLMs to identify performance-improving source code edits for warehouse-scale applications. ACE~\cite{ace} investigates AI-driven optimizations for these environments. Magellan~\cite{magellan} provides an agentic framework to evolve compiler passes using {\em AlphaEvolve}. \ours applies the latter approach to interprocedural code layout.
\section{Conclusion}
\label{sec:conclusion}
In this work, we present \ours, the first framework to successfully apply fine-grained, interprocedural code layout optimization to large binaries, including warehouse-scale applications.  \ours has been built on top of the {\em Propeller} framework without compromising scalability.
By applying the Magellan \cite{magellan} approach of integrating AlphaEvolve and Vizier, our approach autonomously discovers novel layout policies, navigating an exponentially larger search space,  and overcoming the limitations of traditional heuristics.
We have built a highly stable execution environment that reduces run-to-run variance to under 0.05\% that  enables the evolutionary loop to learn directly from precise hardware performance counters rather than relying on approximate static cost models.
Our evaluation demonstrates clear performance gains on real-world, front-end bound workloads.
Most notably, \textit{\ours} achieves a 0.23\% execution improvement on a heavily optimized, warehouse-scale Search service compared to the state-of-the-art Ext-TSP baseline.  Furthermore, it achieves a 1.6\% improvement on a large real-world application, {\em LLVM clang} compiler.
These results confirm that AI-driven agentic workflows can successfully extract additional performance from mature compiler infrastructures.
\section*{Acknowledgments}
We are grateful to Tipp Moseley, James Laudon, Rong Xu, and Teresa Johnson for their valuable feedback on earlier drafts of this paper. 
We also thank Po-Sen Huang, Ng\^{a}n (NV) V\~{u}, and the AlphaEvolve team for their assistance in establishing our experimental framework. 
Finally, we extend our appreciation to the broader team at Google DeepMind for enabling and supporting this research direction.

\bibliographystyle{IEEEtranS}
\bibliography{paper}

\end{document}